\documentclass[journal,twoside,web]{IEEEtran}
\usepackage{generic}
\usepackage{cite}
\usepackage{amsmath,amssymb,amsfonts}

\usepackage{graphicx}
\usepackage{textcomp}

\usepackage{multirow}
\usepackage{comment}
\usepackage{subfigure}

\usepackage{algorithm}
\usepackage{algpseudocode}
\usepackage{algorithmicx}

\makeatletter
\newlength{\trianglerightwidth}
\algnewcommand{\LineComment}[1]{\Statex \hskip\ALG@thistlm $\triangleright$ #1}
\algnewcommand{\LineCommentIndent}[1]{\Statex \hskip2\ALG@thistlm $\triangleright$ #1}
\algnewcommand{\LineCommentCont}[1]{\Statex \hskip\ALG@thistlm%
  \parbox[t]{\dimexpr\linewidth-\ALG@thistlm}{\hangindent=\trianglerightwidth \hangafter=1 \strut$\triangleright$ #1\strut}}
\makeatother

\makeatletter
\newcommand{\algmargin}{\the\ALG@thistlm}
\makeatother
\algnewcommand{\parState}[1]{\State%
  \parbox[t]{\dimexpr\linewidth-\algmargin}{\strut #1\strut}}
  
\algrenewcommand\alglinenumber[1]{\small #1:}

\usepackage[labelfont=bf]{caption} 
\def\BibTeX{{\rm B\kern-.05em{\sc i\kern-.025em b}\kern-.08em
    T\kern-.1667em\lower.7ex\hbox{E}\kern-.125emX}}
\begin{document}
\title{Adversarial Attention for Human Motion Synthesis}
\author{ Matthew Malek-Podjaski, Fani Deligianni \IEEEmembership{Member, IEEE}
\thanks{The authors acknowledge funding from EPSRC EP/W01212X/1 and the Royal Society RGS/R2/212199 }
\thanks{Matthew Malek-Podjaski is with School of Computing Science, University of Glasgow, UK, (e-mail: 2323841m@student.gla.ac.uk). }
\thanks{Fani Deligianni is with the School of Computing Science, University of Glasgow, UK, (e-mail: fani.deligianni@glasgow.ac.uk).}
}

\maketitle

\begin{abstract}
Analysing human motions is a core topic of interest for many disciplines, from Human-Computer Interaction, to entertainment, Virtual Reality and healthcare. Deep learning has achieved impressive results in capturing human pose in real-time. On the other hand, due to high inter-subject variability, human motion analysis models often suffer from not being able to generalise to data from unseen subjects due to very limited specialised datasets available in fields such as healthcare. However, acquiring human motion datasets is highly time-consuming, challenging, and expensive. Hence, human motion synthesis is a crucial research problem within deep learning and computer vision. We present a novel method for controllable human motion synthesis by applying attention-based probabilistic deep adversarial models with end-to-end training. We show that we can generate synthetic human motion over both short- and long-time horizons through the use of adversarial attention. Furthermore, we show that we can improve the classification performance of deep learning models in cases where there is inadequate real data, by supplementing existing datasets with synthetic motions. 
\end{abstract}

\section{Introduction}

Synthesising human motions is an active research problem with many cross-disciplinary applications. Popular research areas often focus on generating human motions for Human-Computer Interaction (HCI) and entertainment applications. These vary from generating gesture motions from speech \cite{alexanderson_gesture_2020} to creating realistic virtual conference avatars to animating video game characters in real-time from ambiguous control signals sent from a video game controller \cite{holden_phase-functioned_2017, zhang_mode-adaptive_2018, starke_neural_2019, starke_local_2020}. Synthetic motions can also be used for modelling realistic human-to-human interactions to improve the realism of Virtual Reality (VR) applications \cite{men_gan-based_2021}. Furthermore, human motions are highly sought after in more niche applications such as simulating crowd movements \cite{guy_pledestrians_2010} or predicting pedestrian motions \cite{mangalam_disentangling_2020}, which are necessary for urban planning, traffic engineering, and self-driving vehicles. 

However, one area of research that has not been explored is synthetic data in human motion analysis problems. Good quality human motion data is challenging to capture because it requires a dedicated space and an expensive professional motion capture setup. This makes deep learning challenging since it depends on the availability of large and balanced datasets. Lack of data often results in models that show promising results on a specific set of data; however fail to generalise to other datasets, especially when the data comes from unseen subjects. Suppose we can train deep learning models to generate realistic human motions. In that case, we could use them to create synthetic motions datasets on which we can train human motion analysis models instead of recording many hours of expensive motion capture data.

Previous techniques have applied deterministic methods, such as recurrent Long Short-Term Memory (LSTM) models \cite{jain_structural-rnn_2016, chiu_action-agnostic_2019, li_bidirectional_2019}, phased-functioned models \cite{holden_phase-functioned_2017}, and Mixture-of-experts models \cite{zhang_mode-adaptive_2018, starke_local_2020}. However, deterministic recurrent models often suffer from averaging poses, where the network's output eventually collapses into a mean pose. On the other hand, phase-functioned networks can generate both short- and long-term motions without error accumulation. However, they require manually designed phase functions that make assumptions about the nature of the motion making them difficult to apply in generalised scenarios. Another concern with many prior techniques is that they mostly rely on subjective observer reports to evaluate the effectiveness of the proposed methods rather than demonstrating comprehensive quantitative results \cite{holden_phase-functioned_2017, alexanderson_gesture_2020, henter_moglow_2020}.

We propose using a probabilistic approach, similar to recent work applying probabilistic Normalizing Flow methods to human motion synthesis \cite{henter_moglow_2020}. We argue that a probabilistic approach can more accurately represent the non-determinism of human motions and generate more accurate and more varied movements. This is beneficial for both HCI applications, where we wish to generate more realistic motions, and synthetic motion datasets, as we can introduce more variety into the data.

We propose a novel probabilistic deep adversarial architecture for learning both short- and long-term motion synthesis.  We use an Attention-based Wasserstein Generative Adversarial Network with Gradient Penalty that we call the Attention WGAN-GP. Our model makes no assumptions about the underlying motion data or the control signals, which we show by training our model to generate various action motions. Through the use of attention and autoregression, we also show that our model can continuously generate realistic motions over both long- and short-time horizons. Our approach outperforms commonly used LSTM models in human motion synthesis. Furthermore, we provide a quantitative approach to evaluating the synthetic motions generated by our model. Through the application of our proposed model as a data augmentation technique in activity recognition, we show that the motions generated by our method perform similarly to real motion capture data and even significantly improve the classification performance when used to generate additional training data in scenarios with inadequate real motion-captured data.

\begin{figure*}[!t]
\begin{center}
\includegraphics[width=1.0\linewidth]{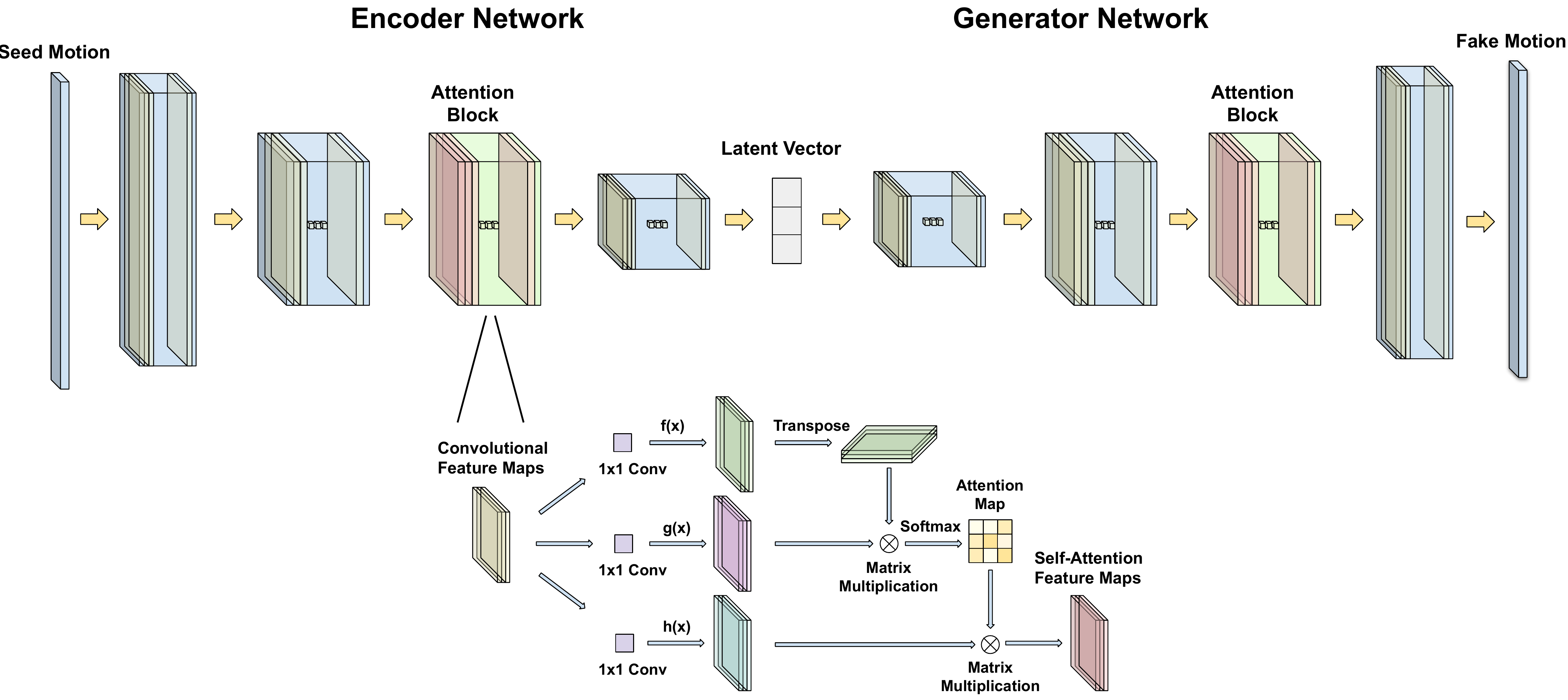}
\end{center}
\caption{The architecture of the Generative network. The seed motion is encoded into a bottleneck latent code, from which the generator network predicts future motions. For the Self-Attention module, the convolutional feature maps are transformed into three separate representations, f, g, and h, where f and g are composed through a matrix multiplication to create the attention map that is applied to the transformed feature map h to generate the self-attention feature maps. The same Self-Attention module is applied to the critic in the second last layer.}\label{fig:generator}
\end{figure*}

\section{Background}
\subsection{Human Motion Analysis and Motion Synthesis}

The analysis of human motions involves extracting the relation between body movements and well-being. Towards this end a large body of research work is concerned about the detection of abnormalities in a person's kinematics \cite{gu_cross-subject_2020,Guo19}, recognise actions \cite{bi_human_2021, li_symbiotic_2019}, or affective states \cite{deligianni_emotions_2019, malek_podjaski_2021}. However, healthcare data is often scarce, and class imbalances are prevalent \cite{frid-adar_synthetic_2018}, making it challenging to deploy generalisable machine learning models. Both shallow \cite{hutchison_gender_2005,li_emotion_2016} and deep learning \cite{ravanbakhsh_plug-and-play_2018}  classifiers are prone to overfitting the training data and often require robust regularisation techniques such as dropout, or specific manual feature extraction, to make the models learn the abstract features of the motions.

Generalisability of the models often improves with access to more data samples from a wider variety of subjects and larger datasets. However, acquiring large human motion data sets is challenging and expensive. Data in healthcare applications are also limited due to ethical and privacy concerns \cite{malek_podjaski_2021, ma_motion_2006, li_emotion_2016, hutchison_gender_2005}. Therefore, synthetic motion data generation could play an important role in human motion analysis models.

The synthesis of human motion largely fit into two categories, offline methods that are designed for full motion prediction \cite{holden_deep_2016}, and online methods designed for continuous motion generation \cite{fragkiadaki_recurrent_2015}. 
Offline methods are used where the motions being generated can easily be segmented, and thus the generative model can predict the motion from start to end. On the other hand, online models generate motions based on small sequences at a time, allowing them to be more flexible in the types of motions they can generate as they can deal with variable length motions. 
To perform online motion prediction, models are typically trained using autoregression, where the model predicts the short-term future poses, from a sequence of past poses. 

\subsection{Recurrent Models}
Recurrent Neural Networks (RNN) are designed to process time-series data sequentially and thus can handle variable-length motions. A typical application of RNNs in motion synthesis is the LSTM architecture. LSTMs have been combined with spatio-temporal graphs \cite{jain_structural-rnn_2016} and hierarchical structures \cite{chiu_action-agnostic_2019} to capture high-level motion information within motion synthesis. They have been used alongside autoencoder models \cite{ghosh_learning_2017, li_bidirectional_2019} and Encoder-Recurrent-Decoder (ERD) models \cite{fragkiadaki_recurrent_2015} to reduce motion drift. They have been also incorporated into probabilistic models to improve the quality of the generated motions \cite{henter_moglow_2020}.

An alternative to the LSTM often applied for human motion synthesis is the Gated Recurrent Unit (GRU). GRUs reduce the number of gates needed by the RNN to avoid problems with vanishing gradients, thus making them more computationally efficient and easier to train \cite{guo_human_2019}. GRUs are often used within encoder and decoder networks \cite{li_dynamic_2020, vedaldi_dlow_2020} to extract motion features and predict future frames in a motion synthesis context. They have been used to forecast context-aware motions, such as modelling motions driven by interactions with other humans or objects \cite{corona_context-aware_2020, adeli_socially_2020}.  
 
However, a challenge in using recurrent models is that errors in the predicted poses are fed back into the input and accumulates \cite{mourot_survey_2022, martinez_human_2017}, causing them to fail in long-term motion prediction. A commonly observed consequence of this is that despite an RNN dominating in short-term prediction \cite{fragkiadaki_recurrent_2015}, the motions eventually collapse to an average pose \cite{fragkiadaki_recurrent_2015, zhang_mode-adaptive_2018, mourot_survey_2022}.

\subsection{Phase Functioned Neural Networks}

The issue of averaged pose generation has been tackled in real-time motion synthesis problems through Phase Functioned Neural Networks (PFNN) \cite{holden_phase-functioned_2017}. These models optimise a phase function of neural network weights rather than a single set of weights, such that a different network setup is responsible for generating each pose of the motion. 

For a given motion, if we can define it in terms of a start and endpoint, we can design a function that describes the different time steps of the motion as a curve. For example, ideally, with a repetitive motion like a gait cycle, we can define the start as a heel strike of the right foot and the end as the next heel strike of the same foot. Meaning that the function through which we describe the motion is cyclical, and we can define the phase of the motion (time step) as the range between 0 and $2\pi$. The outputs of this phase function are a set of trained neural network weights, and each is an expert at generating the pose at a given phase. Using different network weights for each pose, we can avoid the issue of motion blending as this technique effectively trains a separate neural network for each consecutive pose. 

The concept of phase functions was further extended with the introduction of local motion phases \cite{starke_local_2020}. Local motion phases were used in video games to generate basketball motions, where there is a need to independently model motions of different body parts, such as asynchronously animating the legs and the arms. By defining motion phases by contact points between the character's joints and objects in the environment, the motion phase can be broken down from a single global phase function into a set of local phases for each joint, thus allowing for asynchronous animation of each body part.

Phase functions may not generalise well to other less repetitive motion types as they require the segmentation of each motion by start and endpoints, which in practice are difficult to define \cite{zhang_mode-adaptive_2018}. In this case, rather than letting the model map the differences between each motion internally, an external phase function needs to be manually defined for each motion \cite{mourot_survey_2022}.

To address the limitations of the PFNN, a Mixture of Experts (MoE) model was developed \cite{zhang_mode-adaptive_2018} for motion synthesis. MoE models build upon the idea of using different network weights for each phase/motion; however, rather than manually designing a phase function, a gating network is used in its place. This gating network is a neural network responsible for outputting a set of expert model weights used to update the generative model at runtime. A given expert generative model is then responsible for generating the pose at the current time frame. This method has shown success in generating gait motions for quadrupeds \cite{zhang_mode-adaptive_2018}. This was further extended to bipedal characters performing various actions and scene interactions through a goal-driven bi-directional control scheme \cite{starke_neural_2019}. The model seamlessly switches between two levels of control: a high-level goal-driven mode where a user can select an object for the model to interact with and a low-level locomotion mode where the model generates walking animations. The accuracy of the interactions was also greatly improved by synthesising the motions bi-directionally.

Replacing the phase function with a neural network extends the range of motions that can be learned by a single model and even allows for the blending of different expert model weights to create combinations between different motions. However, a potential down side is that we do not have direct control over which aspects of motion are learned by each individual expert network \cite{zhang_mode-adaptive_2018}.

\begin{figure*}[!t]
\begin{center}
\includegraphics[width=0.80\linewidth]{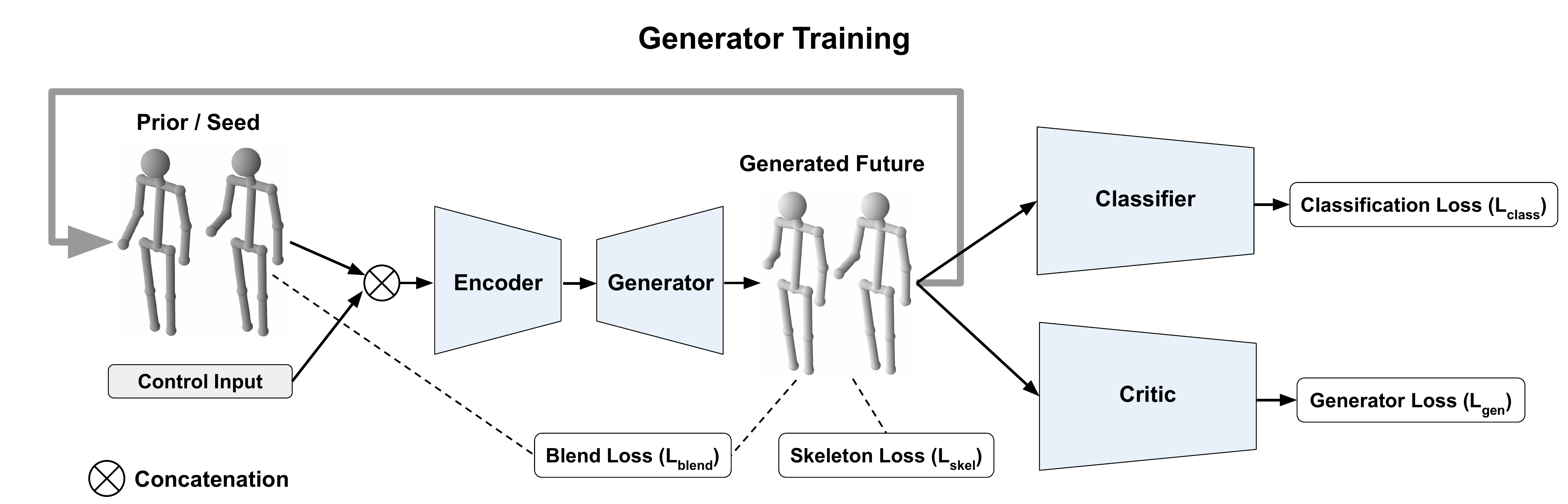}
\end{center}
\caption{Training for the generator network of the GAN model. Prior motions and a control input are fed into the generator model, consisting of an encoder and generator network. The resulting motions are then used to calculate the blending and skeleton losses. The generated motions are also passed through the classifier and critic models to calculate the classification and Wasserstein generator loss. Finally, the generated motion is passed back into the input as a prior motion for the next generator iteration of the autoregressive training.}\label{fig:generator-training}
\end{figure*}

\begin{figure*}[!t]
\begin{center}
\includegraphics[width=0.99\linewidth]{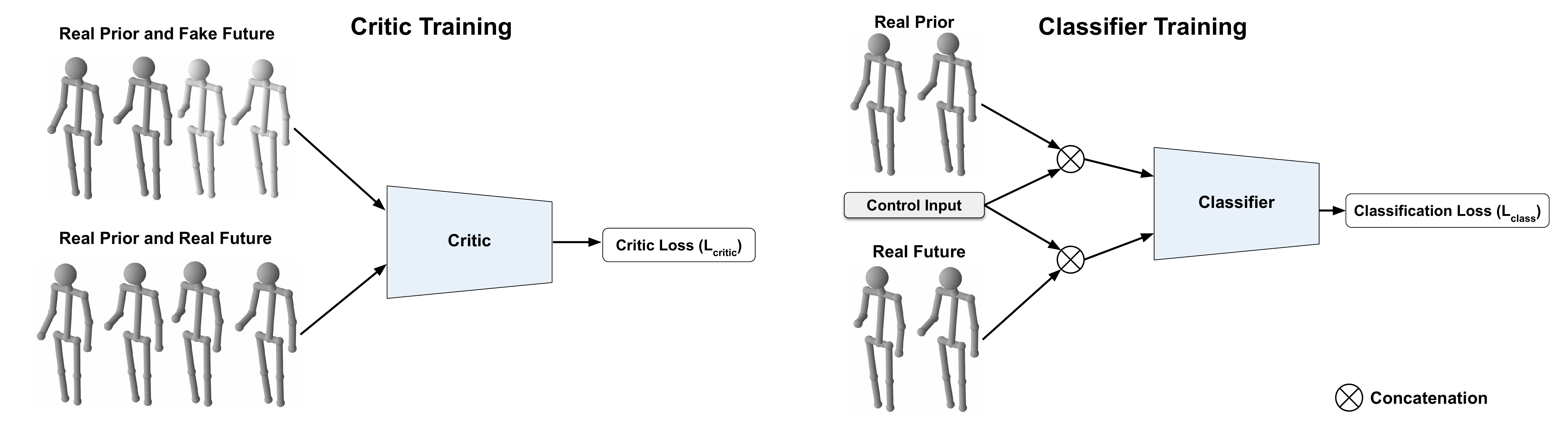}
\end{center}
\caption{Training for the critic and and classifier networks of the GAN model. The critic is trained on both real and generated motions, whereas the classifier is only trained on real data. All the network are trained concurrently in an end-to-end fashion.}\label{fig:critic-training}
\end{figure*}

\subsection{Gaussian Mixture Models and Variational Autoencoders}

Early methods attempted probabilistic motion synthesis using Gaussian Mixture Models (GMM). This allows the development of an online motion generation process that  captures better the non-deterministic nature of human motions. These probabilistic models assume all data is sampled from a mixture of Gaussian distributions with unknown parameters. By learning the parameters for these Gaussian distributions, we can sample new poses to generate motions. GMMs have been applied in motion synthesis in combination with recurrent neural network models; specifically, an LSTM model feeding into a GMM \cite{crnkovic_gaussian_2016}, where the LSTM outputs the parameters of the Gaussian distributions rather than the pose directly. This method alleviates the stagnating output problem of recurrent models. 

Another probabilistic architecture that relies on Gaussian distribution is a Variational Autoencoder (VAE). Autoencoder models are deep neural networks designed to compress input data into a lower-dimensional latent code and then reconstruct the data from that latent code back into the original data. A VAE extends the autoencoder architecture by encoding the data as a normal distribution over the latent space. As a result, the latent space of a VAE represents an approximate posterior distribution of the data, from which samples can be drawn and reconstructed by the decoder model to generate new data.

VAE models have been expanded and applied in human motion synthesis tasks by generating continuous walking motion sequences, one frame at a time, through autoregression and an MoE architecture \cite{ling_character_2020}. VAEs incorporating recurrent LSTM cells into the encoder were also used to generate an autoregressive distribution in the latent space representing a motion canvas from which new motion frames can be interpolated using the decoder \cite{habibie_recurrent_2017}. Motion graph methods \cite{min_motion_2012} have also been used with VAEs to generate locomotion through style transfer \cite{du_stylistic_2019}. A motion graph is a graphical representation of a motion broken down into a sequence of primitive motions, such as lifting a left foot or placing a right foot. A VAE model is used to learn the distribution of motion primitives for each motion. GMMs and VAEs assume that the data is sampled from a mixture of Gaussian distributions, which can limit their ability to simulate subtle motions and variations.

\subsection{Normalizing Flows Networks}
A more recently successful method for generating motions is normalizing flows. This approach designs probability distributions through a series of invertible operations \cite{kobyzev_normalizing_2021}. By applying a series of invertible functions to a complex probability distribution, they transform it into a valid simpler probability distribution. Using invertible operations means that Normalizing Flows performs exact inference, generally producing sharper results as data is sampled directly from the underlying probability distribution without approximation. Normalizing Flows have been explored in models such as MoGlow \cite{henter_moglow_2020} for controllable human locomotion generation and the synthesis of speech-driven virtual character gestures \cite{alexanderson_stylecontrollable_2020}. However, a drawback of normalizing flow models is that they generally require more parameters than models like VAEs or Generative Adversarial Networks (GAN) to model complex probability distributions.

\subsection{Generative Adversarial Networks}

Generative Adversarial Networks (GANs) are generative models that learn complex probability distributions through adversarial learning. Rather than minimising the direct motion reconstruction error, a generative network is trained to fool a discriminator network. The discriminator predicts whether its input is a real data sample or a synthetic created by the generative network. During training, the discriminator minimises its prediction loss to classify real and fake samples correctly. In contrast, the generative network aims to maximise the prediction loss in an adversarial fashion. This approach to generator training is powerful because rather than learning to recreate the specific data samples in a given dataset, the generative network learns to generate data that could plausibly have been sampled from the true data population. Although any motion generation network can be wrapped into a GAN by incorporating a discriminator into the training \cite{mourot_survey_2022}, a popular architecture for GANs is based on adversarial autoencoders \cite{makhzani_adversarial_2016}. The motion is first encoded into a latent code, from which future motions are predicted using the decoder.

Adversarial models have been applied in various ways for human motion synthesis. By using random latent codes drawn from a Gaussian process, a generator network has been used to gradually increase the resolution in the spatial and temporal dimensions through graph upsampling operations to build up a sequence of human skeletons \cite{yan_convolutional_2019}. Sequence-to-sequence (Seq2Seq) GAN architectures were applied to predict future motions in offline applications \cite{barsoum_2018}. A Seq2Seq model is an architecture where a convolutional encoder is used to transform the motion into a latent code from which a generator can predict future motion frames. Adversarial models have also been used to develop motion style transfer techniques where two decoders are trained, one for creating a latent code for style, and another for a latent code of the motion. Subsequently, a decoder is used to generate a motion that blends the given style and motion \cite{wang_adversarial_2020}.

GANs have also been used with RNNs to predict motion frames between sparse key animation frames in a process similar to in-betweening in traditional character animation \cite{harvey_robust_2020} or to refine motions produced by an LSTM \cite{wang_combining_2018}. Human to human motion interactions have also been modelled using recurrent adversarial autoencoders \cite{men_gan-based_2021}.

\begin{algorithm}[ht]
\small
\caption{\small WGAN with gradient penalty and classifier. We use default values of $\gamma = 10$, $n_{critic} = 6$, $n_{generator} = 2$, $\alpha = 0.005$, $\beta_{1}  = 0$, $\beta_{2} = 0.9$}\label{alg:wgan}
\begin{algorithmic}[1]
\Require The gradient penalty coefficient $\gamma$, number of critic iterations $n_{critic}$, number of generator iterations $n_{generator}$, batch size $m$, Adam hyperparameters $\alpha$, $\beta_{1}$, $\beta_{2}$.
\Require initial critic parameters $w_{0}$, initial classifier parameters $v_{0}$, initial generator parameters $\theta_{0}$.
\While{$\theta$ has not converged}
    \For{$t = 1$, ..., $n_{critic}$}
        \LineCommentIndent Critic step
        \For{$i = 1$, ..., $m$}
            \parState{Sample real data $\boldsymbol{x}, \boldsymbol{z}$, and a random number $\epsilon \sim U[0, 1].$}
            \State{$\boldsymbol{\tilde{x}} \gets G_{\theta}(\boldsymbol{x})$}
            \State{$\boldsymbol{\hat{x}} \gets \epsilon+(1-\epsilon)\tilde{\boldsymbol{x}}$}
            \State{$\boldsymbol{r} \gets (\boldsymbol{x},\boldsymbol{z})$}
            \State{$\boldsymbol{f} \gets (\boldsymbol{x},\tilde{\boldsymbol{x}})$}
            \State{$L_{gp}^{(i)} \gets \left(\left\|\nabla_{\hat{\boldsymbol{x}}}D_{w}(\hat{\boldsymbol{x}})\right\|_{2}-1\right)^{2}$}
            \State{$L_{critic}^{(i)} \gets D_{w}(\boldsymbol{r})-D_{w}(\boldsymbol{f})+\lambda L_{gp}^{(i)}$}
        \EndFor
        \State{$w \gets \operatorname{Adam}\left(\nabla_{w} \frac{1}{m} \sum_{i=1}^{m} L_{critic}^{(i)}, w, \alpha, \beta_{1}, \beta_{2}\right)$}
        \LineComment Classifier step
        \State{Sample a batch of real data $\left\{\boldsymbol{x}^{(i)}, \boldsymbol{z}^{(i)}, \boldsymbol{y}^{(i)}\right\}_{i=1}^{m}$.}
        \State{$\boldsymbol{\tilde{y}} \gets C_{v}(\boldsymbol{z})$}
        \State{$L_{class} \gets$ $-\frac{1}{m} \sum_{i=1}^{m} y_{i} \cdot \log \left(\hat{y}_{i}\right)$}
        \State{$v \gets \operatorname{Adam}\left(\nabla_{v} \frac{1}{m} \sum_{i=1}^{m}L_{class}, \theta, \alpha, \beta_{1}, \beta_{2}\right)$}
    \EndFor
    \LineComment Generator step
    \State{Sample a batch of real data $\left\{\boldsymbol{x}^{(i)}, \boldsymbol{z}^{(i)}, \boldsymbol{y}^{(i)}\right\}_{i=1}^{m}$.}
    \State{$\tilde{x} \gets G_{\theta}(x)$}
    \For{$k = 1$, ..., $n_{generator}$}
    \State{$\boldsymbol{\hat{x}} \gets G_{\theta}(\tilde{x})$}
    \State{$\boldsymbol{\hat{y}} \gets C_{v}(\boldsymbol{\hat{x}})$}
    \State{$L_{class} \gets$ $-\frac{1}{m} \sum_{i=1}^{m} y_{i} \cdot \log \left(\hat{y}_{i}\right)$}
    \State{$L_{skel} \gets \frac{1}{T}\sum\limits_{{\mathop{\rm s}\nolimits}  \in S} {\sum\limits_{t = 0}^T {\left\| {\boldsymbol{x}_0^s - \hat{\boldsymbol{x}}_t^s} \right\|}}$}
    \State{$L_{blend} \gets \mathbb{E}\left[ {{{\left( {\left\| {{\tilde{\boldsymbol{x}}_T} - {\hat{\boldsymbol{x}}_0}} \right\|} \right)}^2}} \right]$}
    \State{$L_{gen} \gets -\mathbb{E}\left[D_{w}(\hat{x})\right] + L_{blend} + L_{skel} + L_{class}$}
    \State{$\theta \gets \operatorname{Adam}\left(\nabla_{\theta} \frac{1}{m} \sum_{i=1}^{m}L_{gen}, \theta, \alpha, \beta_{1}, \beta_{2}\right)$}
    \State{$\tilde{x} \gets \hat{x}$}
    \EndFor
\EndWhile
\end{algorithmic}
\end{algorithm}

\section{Methods}
\begin{figure*}[!h]
\begin{center}
\includegraphics[width=0.8\linewidth]{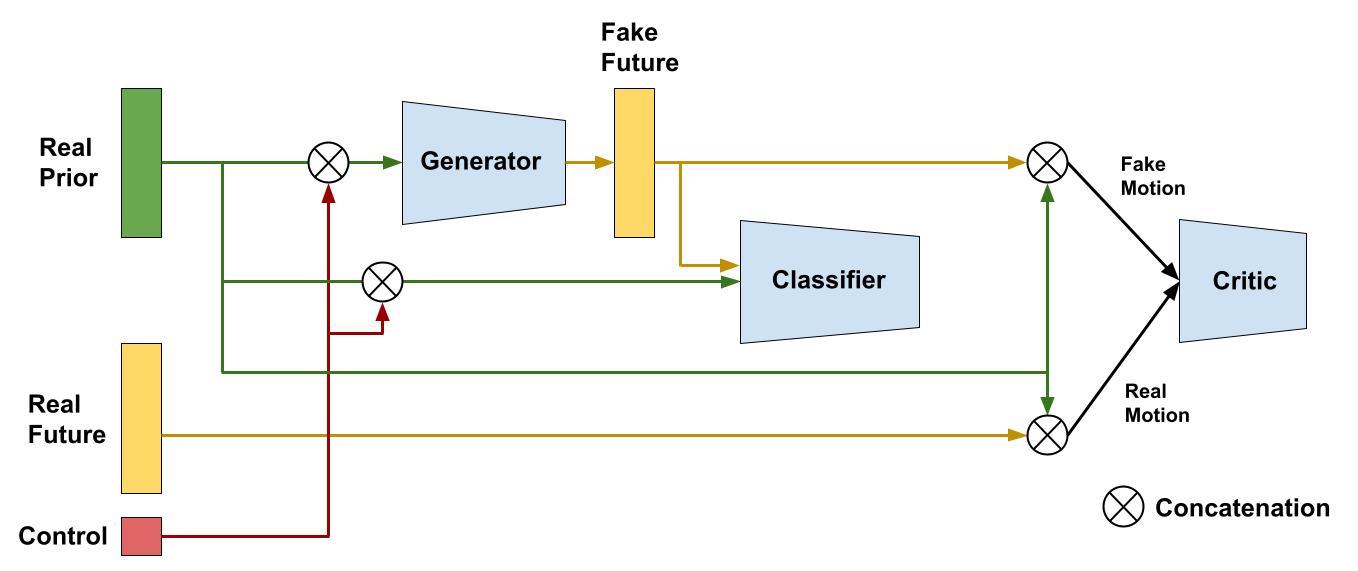}
\end{center}
\caption{Visualisation of the end-to-end training of the Attention WGAN-GP architecture.}\label{fig:end_to_end}
\end{figure*}

To generate synthetic motions, we propose a novel online generative adversarial probabilistic model. Specifically, we exploit a Wasserstein Generative Adversarial Network (WGAN) \cite{arjovsky_wasserstein_2017} with Gradient Penalty (WGAN-GP) \cite{gulrajani_improved_2017} and Self-Attention \cite{zhang_self-attention_2019} modules for improved synthetic motion accuracy, in combination with a classifier network \cite{lee_controllable_2019} for controlling the motion generation. Our architecture is summarised in Figure \ref{fig:generator}, and from here on we refer the model as the Attention WGAN-GP. An overview of the end-to-end training is presented in \ref{fig:end_to_end}. The training of each network in the Attention WGAN-GP architecture is outlined visually in Figures \ref{fig:generator-training} and \ref{fig:critic-training}. A more detailed training algorithm is also provided in Algorithm \ref{alg:wgan}, explained in sections \ref{sec:training} and \ref{sec:loss}. 

The WGAN model is an extension of the standard GAN, which improves the stability of the training \cite{arjovsky_wasserstein_2017}. In this case the network that determines whether a motion is real or fake is referred as a critic rather than a 'discriminator'. A traditional discriminator outputs a probability of whether a given motion is real or fake, whereas a critic scores the motion on how realistic it believes it is. This allows for stable training of an adversarial model with motion data and prevents the discriminator network to converge quickly, leaving no room for the generator to improve. The critic converges to a linear function which allows for the calculation of gradients in all states \cite{arjovsky_wasserstein_2017}, thus allowing the generator to converge to optimality.

Many adversarial motion synthesis models generate data from random noise vectors. However, the generation process needs to be controllable for synthetic motions to supplement existing datasets. For controllable adversarial motion generation, we need to turn to models such as Conditional GANs \cite{mirza_conditional_2014} or ControlGANs \cite{lee_controllable_2019}. A Conditional GAN conditions the disciminator on the real or fake label alongside a class label. Here we adapted a ControlGAN approach, which separates the prediction of each label into individual networks. 

\subsection{Self-Attention}
Attention mechanisms are applied in GANs through a self-attention module \cite{zhang_self-attention_2019} to guide the model to relate distant data sections, such as distant poses in a motion. 
Although attention has been recently explored within motion synthesis applications \cite{valle-perez_transflower_2021, aksan_spatio-temporal_2021}, to our knowledge, it has not yet been used within an adversarial scenario. Since the task of generating motions inherently relies on the spatio-temporal relation of past and future motions, we can leverage attention mechanisms to reduce error accumulation in autoregressive models \cite{aksan_spatio-temporal_2021} for more accurate short- and long-term motion prediction. 

Self-attention works by extending the convolutional layer of the GAN with an additional term, the attention map, which is added to the output. The attention map acts as a mask that determines the contribution that a particular section of the data sample has on the generation of another section. Initially, the model will explore only the local area of each section as it performs the convolution, just as a typical convolutional GAN. However, over time, it can learn how different output regions relate to each other and condition the generated output on related areas.
Self-attention is applied to both the generator and critic networks, allowing both to learn distant spatial relationships in the data, resulting in more realistic motion synthesis.


According to the original definition for the Self-Attention GAN module \cite{zhang_self-attention_2019}, the features from the previous layer $\boldsymbol{x} \in$ $\mathbb{R}^{C \times N}$ ($C$ being the number of channels and $N$ the number of feature locations) are transformed into two feature spaces $\boldsymbol{f}, \boldsymbol{g}$ to calculate the attention, where $f(x)=W_{f} x, g(x)=$ $\boldsymbol{W}_{\boldsymbol{g}} \boldsymbol{x}$

\begin{equation}
\beta_{j, i}=\frac{\exp \left(s_{i j}\right)}{\sum_{i=1}^{N} \exp \left(s_{i j}\right)}, \text { where } s_{i j}=\boldsymbol{f}\left(\boldsymbol{x}_{\boldsymbol{i}}\right)^{T} \boldsymbol{g}\left(\boldsymbol{x}_{\boldsymbol{j}}\right)
\end{equation}

The attention map $\beta_{j, i}$ creates a mapping between the $i^{th}$ and $j^{th}$ regions, which is the amount the model attributes to the $i^{th}$ region, when generating $j^{th}$ region. Thus, the output of the attention layer is $\boldsymbol{o}=\left(\boldsymbol{o}_{1}, \boldsymbol{o}_{2}, \ldots, \boldsymbol{o}_{\boldsymbol{j}}, \ldots, \boldsymbol{o}_{N}\right) \in$ $\mathbb{R}^{C \times N}$, where,

\begin{equation}
\boldsymbol{o}_{\boldsymbol{j}}=\boldsymbol{v}\left(\sum_{i=1}^{N} \beta_{j, i} \boldsymbol{h}\left(\boldsymbol{x}_{\boldsymbol{i}}\right)\right), \boldsymbol{h}\left(\boldsymbol{x}_{\boldsymbol{i}}\right)=\boldsymbol{W}_{\boldsymbol{h}} \boldsymbol{x}_{\boldsymbol{i}}, \boldsymbol{v}\left(\boldsymbol{x}_{\boldsymbol{i}}\right)=\boldsymbol{W}_{\boldsymbol{v}} \boldsymbol{x}_{\boldsymbol{i}}
\end{equation}

and $\boldsymbol{W}_{g} \in \mathbb{R}^{\bar{C} \times C}, \boldsymbol{W}_{f} \in \mathbb{R}^{\bar{C} \times C}$, $\boldsymbol{W}_{\boldsymbol{h}} \in \mathbb{R}^{\bar{C} \times C}$, $\boldsymbol{W}_{\boldsymbol{v}} \in \mathbb{R}^{C \times \bar{C}}$ are the weight matrices learned through 1x1 convolutions as seen in Figure \ref{fig:generator}. Finally, the attention map is then added back onto the input feature map multiplied by the scalar parameter $\gamma$.

\begin{equation}
\boldsymbol{y}_{i}=\gamma \boldsymbol{o}_{i}+\boldsymbol{x}_{i}
\end{equation}

The scalar parameter $\gamma$ is initialised to 0 and is used to control the impact of the attention map over time. It allows the mode to begin learning through the exploration of features within the local space, just as a traditional convolutional GAN, and gradually over time learn to give more weight to the non-local features.

\subsection{Autoregression}
A core component of our training architecture is the iterative autoregressive training of the generator to enable it to perform long-term motion generation, as shown in Figure \ref{fig:generator-training}. The motions used for training the generative online model are windowed by one second time intervals. In contrast, for the motion classification task, we need full-length motions that can be used to supplement our existing motion dataset. Furthermore, since the motions in our dataset are 3-4 seconds long, the model needs to accurately predict more than just the next second of motion without compromising its online ability to generalise to varying lengths of motion. Hence, we needed to introduce the multiple critic iterations into the training algorithm.

% ...

\subsection{Model Formulation}
Given a motion $M^{2T}$, of temporal length $2T$, we can define the motion as a sequence of poses $M^{2T} = {m_{0:2T}}=[{m_0},...,{m_{2T}}]$, where each pose $m_{t} \in \mathbb{R}^{3J}$, is a skeleton with J being the number of joints, each specified by $\mathnormal{(x\hat{i},y\hat{j},z\hat{k})}$ coordinates, and can be defined as a probability distribution conditioned on the $\tau$ previous poses:

\begin{equation}
p(m_{0:2T}) = \prod\limits_{t = \tau}^{2T} {p({m_t}|{m_{t-\tau : t-1}})}
\end{equation}

To make the sampling of poses controllable we also condition the pose probability distribution on an additional parameter $Y$ that acts as a control signal to confine to set of possible poses at each time frame $y_{t} \in \mathbb{R}^{Y}$, similar to other probabilistic motion models \cite{henter_moglow_2020}. This gives us a probability distribution for each pose conditioned on the set of previous poses and a control vector y:

\begin{equation}
p(m|y) = \prod\limits_{t = \tau}^{2T} {p({m_t}|{m_{t-\tau : t-1}}, y_{t-\tau:t})}
\end{equation}

Splitting the motions into two sequences of length $T$ we can define a set of prior motions $P$ (also referred to as seed motions) and a set of future poses $F$, where $x=m_{0:T-1}$, $z=m_{T:2T}$ and $(x \in P)$, $(z \in F)$.

Our aim to create an architecture where we can model the underlying probability distribution that maps a sequence of prior poses to a sequence of future poses. Towards this end a generator $G$ is trained to generate synthetic sequences of future poses from sequences of prior poses, $\tilde{z} = G(x)$.

By recursively feeding the generated poses back into the generator network we can continuously generate motions autoregressively, given we have an initial seed motion.

\subsection{Loss Functions}
\label{sec:loss}
To train our model, we use five different objective functions. A critic loss function based on the gradient penalty Wasserstein critic loss \cite{gulrajani_improved_2017} ($L_{critic}$), a cross-entropy classifier loss ($L_{class}$), and a combination of three loss functions for the generator ($L_{gen}$). $L_{gen}$ consists of a generator Wasserstein loss \cite{arjovsky_wasserstein_2017}, a skeleton loss to constrain the size of the skeleton ($L_{skel}$) and a blending loss ($L_{blend}$) to improve continuity between prior and future motions.

\subsubsection{Critic Gradient Penalty Loss}
The critic uses the improved gradient penalty Wasserstein loss \cite{gulrajani_wgan_gp}. The Wasserstein loss requires a Lipschitz constraint on the critic, and the original loss achieves this by applying weight clipping on the critic \cite{arjovsky_wgan}. We found in our training that enforcing a compact space on the weights of the critic network results in a generator network that is unable to output a human pose. Applying a penalty on the gradient norm rather than clipping the weights solves this issue.

Given a real prior motion $x$, a corresponding real future motion $z$, and a generated future motion $\boldsymbol{\tilde{z}} = G(\boldsymbol{x})$. We can define a real motion as $r = (x,z)$, and a fake motion as $f = (x,\tilde{z})$. We define the critic loss as:

\begin{equation}
\label{eq:critic_loss}
L_{critic}={\mathbb{E}}[D(\boldsymbol{r})] - {\mathbb{E}}[D(\boldsymbol{f})]+\lambda L_{gp}.
\end{equation}

To calculate the gradient penalty we need to interpolate between the real and fake motions. As such, with random number sampled from a uniform distribution $\epsilon \sim U[0, 1]$, the interpolated motion is defined as $\boldsymbol{\hat{m}} = \epsilon \boldsymbol{r}+(1-\epsilon)\tilde{\boldsymbol{f}}$. Using the interpolated motion, the gradient penalty is given by:

\begin{equation}
L_{gp}={\mathbb{E}}\left[\left(\left\|\nabla_{\hat{\boldsymbol{m}}} D(\hat{\boldsymbol{m}})\right\|_{2}-1\right)^{2}\right].
\end{equation}

\subsubsection{Generator Loss}
For the generator loss we make use of the original generator Wasserstein loss function \cite{arjovsky_wgan}, which is the negative of the critic output for the fake generated motions. 

\begin{equation}
\label{eq:generator}
L_{gen}=-{\mathbb{E}}[D(\boldsymbol{f}))] + L_{skel} + L_{blend} + L_{class}
\end{equation}

We also extend the generator loss with a skeleton loss, a blend loss and a classification loss.

\subsubsection{Skeleton Loss}
A skeleton loss enforces a physics constraint on the shape of the skeleton to avoid it changing shape during the generated motion. Without constraining the size of the skeleton, the bone lengths in the skeleton would often vary throughout the motion, something that is not physically possible \cite{li_perceptual-based_2020, li_bidirectional_2019, barsoum_2018}. To constrain the bone-length of the skeleton and keep it consistent between prior and future motions, we take the first frame of the prior motion and use its pose as a reference skeleton. Then the skeleton loss is defined as the squared distance between the reference skeleton and the current time frame for a set of joint pairs that define the skeletal bones. 

Given a set of joint pair definitions S that describe a given skeleton. A bone is defined as $s = \{ i,j\} $, $(s\in S)$, where $i$ and $j$ are joints defined by $(x,y,z)$, then $m_t^s = m_t^j - m_t^i$ describes a bone vector of pose $m$ at time $t$. Using the first pose of the prior motion $x_{0}$ as a reference skeleton, we define the skeleton loss of the future motion $z$ as:

\begin{equation}
\label{eq:skeleton}
{L_{skel}} = \frac{1}{T}\sum\limits_{{\mathop{\rm s}\nolimits}  \in S} {\sum\limits_{t = 0}^T {\left\| {\boldsymbol{x}_0^s - \boldsymbol{z}_t^s} \right\|}}
\end{equation}

\subsubsection{Blending Loss}
Another issue observed with only using the Wasserstein loss functions is that the resulting future motions would not continue from past seed motions. As a result, when combining past and future motions, the person would appear to teleport mid-motion. Furthermore, this discontinuity would also results in positional error accumulation in long-term motion generation. Although combining the past and future frames before passing them to the critic alleviated some of these problems, it did not solve them as we had anticipated. We hoped that by giving the critic information from both the past and future motions, it would be able to discern fake motions based on the lack of continuity; however, in practice, this did not seem to be the case.

We found that it was necessary to enforce an additional constraint on the generation to ensure continuity in the motions, the so called blending loss. We calculate the loss to be the distance between the last input motion frame and the resulting first generated motion frame to ensure the generated motions blend with the input motion. This constraint results in synthetic motion that correctly preserve the continuity between past and future motions.

Given a prior motion $(x \in P^(T))$ and a future motion $(z \in F^(T))$, of length $T$, we define the blend loss as the mean square distance between the last prior pose and the first future pose:

\begin{equation}
\label{eq:blending}
{L_{blend}} = \mathbb{E}\left[ {{{\left( {\left\| {{\boldsymbol{x}_T} - {\boldsymbol{z}_0}} \right\|} \right)}^2}} \right]
\end{equation}

\subsubsection{Classification Loss}
For the classifier in our model we use a cross-entropy loss function. We encode the control input as a one-hot encoded vector and use it as the target label for the classifier network.

\begin{equation}
\label{eq:classification}
L_{class}=-\frac{1}{m} \sum_{i=1}^{m} y_{i} \cdot \log \left(\hat{y}_{i}\right)
\end{equation}

The classifier is only ever updated based on the real data during its update. However, during the generator update the classifier weights are frozen and the classification loss is calculated on the fake generated motion. The classification loss is used as part of the generator loss to encourage the generator to produce motions that respect the control signal.

\subsection{Training}
\label{sec:training}
Our training algorithm is specified in Algorithm \ref{alg:wgan}. Much like a typical WGAN, the training of our model is split into multiple stages. Furthermore, we introduce a classifier update stage in addition to the critic and generator update stages. In the original definition of the WGAN \cite{arjovsky_wasserstein_2017} the authors recommend having multiple critic updates for each generator update as the more the critic is trained, the more reliable the Wasserstein loss becomes. Similarly in our model, we perform multiple updates of both the critic network, as well as the model's classification network. 

Firstly for the critic update, the generator is given a 'fake' seed motion from which it generates a fake future motion. Similarly, a 'real' seed motion and the ground truth future motions are concatenated to generate a real motion. A seed motion in this scenario results from the concatenation of a prior motion with the corresponding control vector. The critic loss is then calculated based on how well it scored motions. Furthermore, the critic scores motions randomly interpolated between the real and fake data. The score of which is then used to calculate the gradient penalty ($L_{gp}$), thus giving us the critic loss from Equation \ref{eq:critic_loss}. 

For each critic iteration, the classifier is also updated and it is ensured that it is only trained on real motions. Given a motion, the classifier's task is to predict the corresponding control vector for that motion correctly. This update process is repeated $n_{critic}$ times to ensure that both the critic and classifier have had sufficient time to update before updating the generator.

For updating the generator, firstly, a seed motion is fed through the generator network to create a fake future motion for initialisation. Then in the generator main update step, the fake future is instead used as the seed motion, and the gradients on the loss are calculated on this subsequent consecutive generated motion. This training approach vastly improves the generator model's ability to perform long-term motion prediction. 

The generator loss (Eq. \ref{eq:generator}), the skeleton (Eq. \ref{eq:skeleton}), blending (Eq. \ref{eq:blending}), and classification (Eq. \ref{eq:classification}) losses are calculated on the generated future motion, and this process is repeated for $n_{generator}$ iterations. It is important to note that although all the models are involved in calculating the generator loss, the model weights are only ever updated in their respective update steps. In every other step, the weights of the networks are frozen.

Instead of batch normalisation, we use layer normalisation within our networks to stabilise the model training similar to \cite{gulrajani_wgan_gp}. This also includes replacing batch normalisation layers with layer normalisation in the self-attention modules. 
 This is because the gradient penalty needs to be independently imposed on different samples, whereas batch normalisation brings batch correlation into the model training \cite{salimans_improved_2016}.

\begin{figure*}[!t]
\begin{minipage}{0.96\textwidth}
\begin{center}
\includegraphics[width=0.96\linewidth]{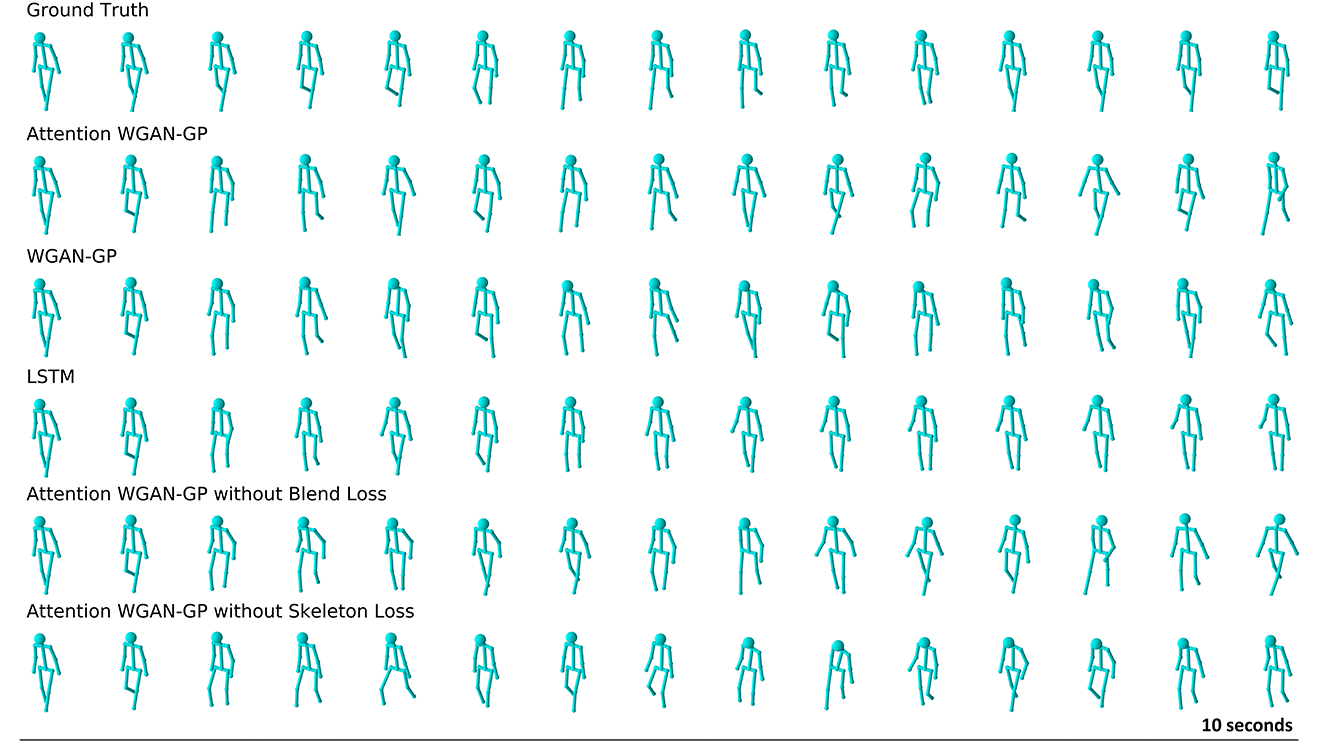}
\end{center}

\begin{center}
\includegraphics[width=0.96\linewidth]{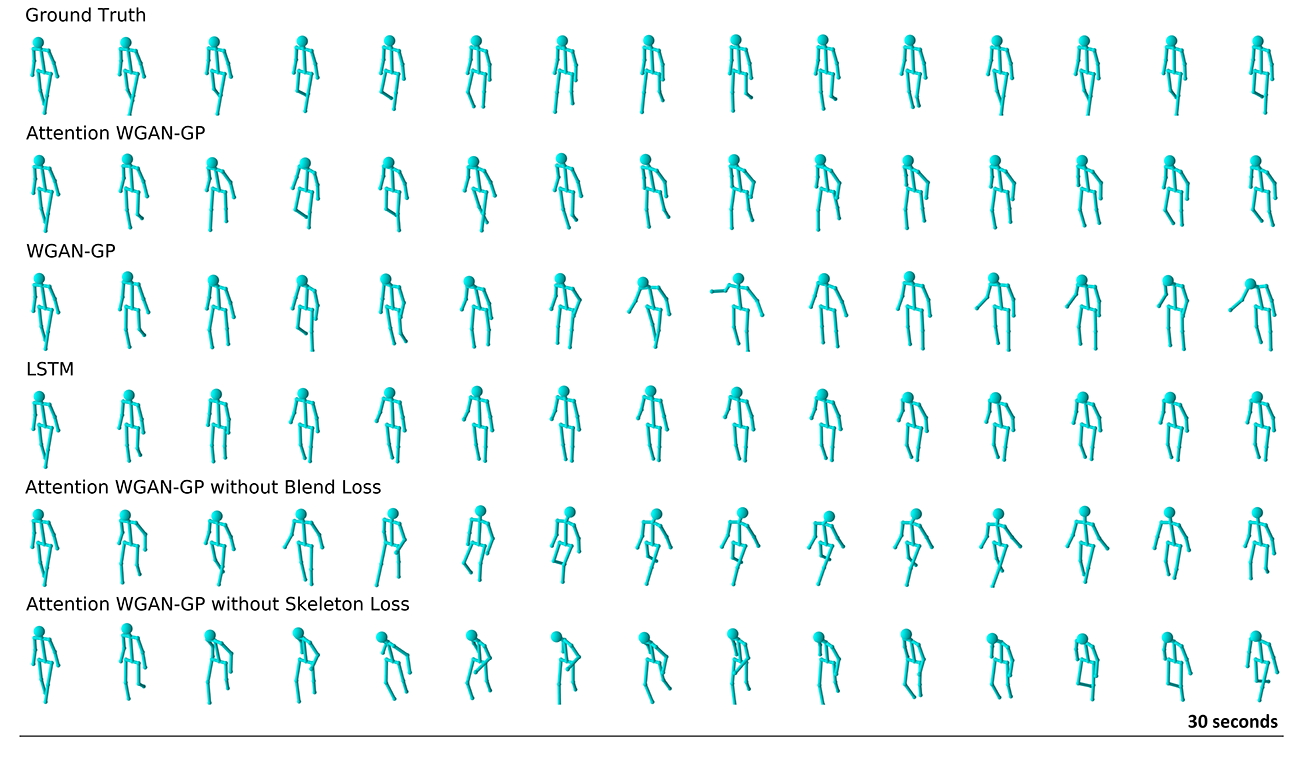}
\end{center}
\caption{A sequence of poses from a 10 and 30 seconds generated motion, respectively. Ground truth motion is compared tothe proposed architecture (Attention WGAN-GP) as well as WGAN-GP and LSTM. Furthermore, we demonstrate the effectiveness of incorporating a blend loss and a skeleton loss. The Attention WGAN-GP model is the only model capable of generating realistic motions over long time sequences.}\label{fig:motion_30s}
\end{minipage}
\end{figure*}

\section{Evaluation Methodology and Results}

\subsection{Dataset} 
To test our models, we use a publicly available motion capture library \cite{ma_motion_2006}. The dataset contains a set of motion captured movements of 27 non-professional subjects (13 male, 14 female, mean age 22, ranging from 17 to 29 years) performing various motions such as walking or throwing, labelled based on their identity, gender and emotion. The available motions in the dataset are: knocking, lifting, throwing, and walking with 64, 88, 64, and 80 motion samples per subject for each action after windowing. The motions were captured using retroreflective markers and a state-of-the-art motion capture system. The raw motion capture data was post-processed through a 3D animation software where the key joint positions from the motion data were projected to outline the human skeleton.

\subsection{Preprocessing}
For preprocessing the data, we centre-mean unit-variance normalise the dataset and remove the effects of global displacement and rotation from the data, similarly to our previous work \cite{malek_podjaski_2021}. This improves the convergence of the model during training. Since the original motion data involves the subjects moving around a room as they perform the actions, the effects of global displacement way out-weight any motion effects. Especially after centre-mean and unit-variance normalisation, the magnitude of the joint movements caused by the action being performed ends up far out-weighed by the extent of the global displacement in the (x,y,z) axes. This makes it challenging for classification models to predict the required motion labels. Therefore, we found much better convergence when training the models by normalising the global displacement. 

\subsection{Qualitative and Quantitative Evaluation}

We train our model to generate various types of motions by using the action labels as control inputs. 
For the evaluation, we want to determine the quality of the generated motions. We present both qualitative results of continuous gait motion as well as quantitative results based on the classification performance of deep learning models trained with and without the synthetic motions generated.
Towards this end a CNN classifier is trained to perform action recognition. We device both an intra-subject stratified K-Fold experiment as well as a leave-one-subject-out (LOSO) cross-validation protocol.  

The walking data used in the experiments is segmented into gait cycles, defined as two consecutive heel strikes of the same foot, interpolated into 125 frames. Likewise, the action data is windowed into 125 frame segments. The synthetic data is generated using the first 25 frames of the motions as the seed, and with 4 generator iterations, a full motion of 125 frames is predicted. From those we drop the initial seed motion frames to ensure only synthetic data is used for the evaluation. Similarly, the ground truth data also only uses the last 100 frames of the motions to ensure the consistent temporal duration of motions between real and synthetic data. 
For each scenario where synthetic data is used for classification, we compare a baseline LSTM model with our Attention WGAN-GP model, using the same input and output data. The baseline LSTM model is trained by directly minimising the mean square error (MSE) between the real future motions and the generated fake future motions. For each motion in the training dataset we generate a corresponding synthetic motion, thus doubling the training data.

\subsection{Results}
Qualitative results are shown in Figure \ref{fig:motion_30s} with sequences of generated poses over 10 and 30 second intervals, respectively. We evaluate the contribution of each feature of the Attention WGAN-GP model by quantifying the quality of the synthetic motions, comparing them to the ground truth data within angular space. We evaluate motions generated by the WGAN-GP model without attention but with both the skeleton loss ($L_{skel}$) and the blending loss ($L_{blend}$). We also evaluate WGAN-GP models with attention but without the skeleton and blending losses. Finally, we evaluate the quality of the generated motions of the Attention WGAN-GP model against a baseline LSTM model. The results show clearly that only the proposed Attention WGAN-GP model is capable of realistic synthetic motion generation over long time horizons. We also note how the quality of the motions degrades over time as different components such as attention or the blend/skeleton losses are removed from the model.

\begin{figure}[!t]
\centering
\hspace{0.05\textwidth}
\begin{minipage}{.472\textwidth}
  \centering
  \includegraphics[width=0.96\linewidth]{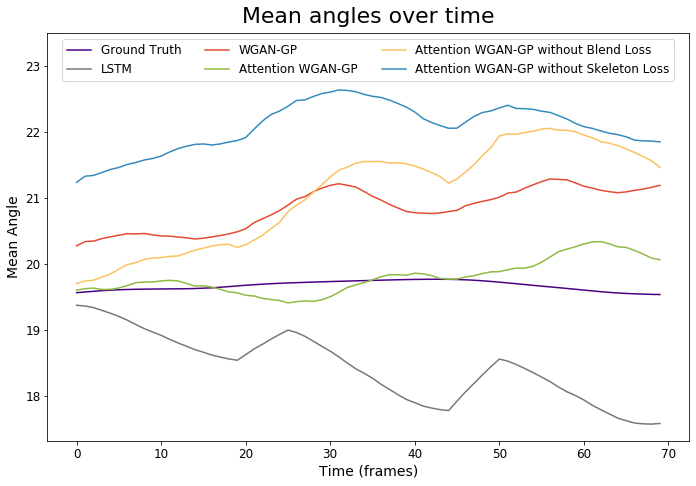}
  \vspace{0.8em}
  \captionof{figure}{Mean angle error over the generation of 70 frames (2.33 seconds) of action motions.}
  \label{fig:mae}
\end{minipage}
\hspace{0.01\textwidth}
\begin{minipage}{.472\textwidth} 
    \centering

    \includegraphics[width=0.96\linewidth]{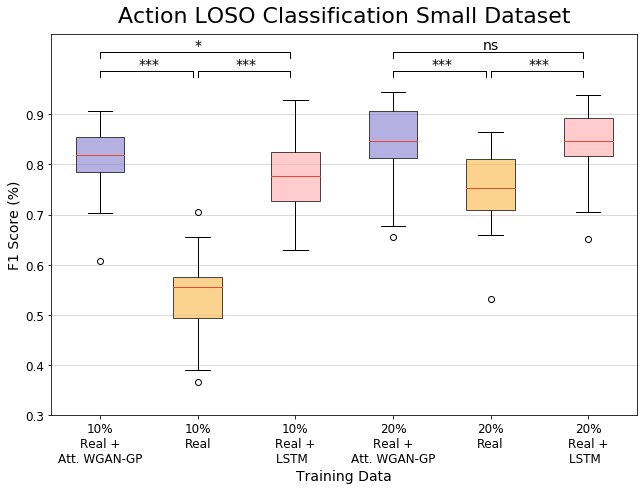}

    \caption{Classification F1 scores for each subject in the LOSO protocol for action classification in a simulated small dataset scenario from experiment 2. Trained on only 10\% and 20\% of the available training data in each fold. We observe that using motion synthesis as a data augmentation technique for increasing the amount of training data can significantly improve classification performance.}\label{fig:small_dataset_loso}

\end{minipage}
\end{figure}

These results are also supported by estimating the angle-space representation, which is a common approach in evaluating synthetic motions \cite{martinez_human_2017, shu_spatiotemporal_2021}. In this case, motions are expressed in a scale and rotation invariant notation by calculating joint angles. This allows a rough quantitative comparison of different approaches with relation to the ground truth. 
Figure \ref{fig:mae} shows the mean angle in the motion data over time for just over 2 seconds of generated motion (motions are recorded at 30 frames per second). We see a significant improvement in the accuracy of the generated motions over time with the addition of attention to the WGAN-GP model. We observe that although the LSTM starts very close to the ground truth data, it quickly collapses into a mean pose, and the mean angles decrease over time. 

Similarly to Figure \ref{fig:motion_30s}, the mean angular estimation in Figure \ref{fig:mae} shows the impact of the blend loss $(L_{blend})$ and the skeleton loss $L_{skeleton}$ in the Attention WGAN-GP model. We observe that with the skeleton loss and without the blend loss, the motion begins close to the ground truth, but it quickly diverges. We have observed that without the blend loss the model does not respect the continuity between the prior and future motions. For example, this may result in the model producing motions that are out of sync by a few frames, and at other times it may jump from walking forward with the left leg to walking forward with the right leg instead. Hence over time, the motions generated from the model without the blend loss quickly deviate from the ground truth motions. 
On the other hand, without the skeleton loss, the model does not have the intrinsic correlation between different joints moving together, and as a result, the accuracy of the generated motions dramatically suffers. 

To evaluate the ability of the synthetic data to improve action classification when available data are limited we simulate a scenario where little motion data is available. We performed an inter-subject, Leave-One-Subject-Out (LOSO) cross-validation evaluation protocol. In each evaluation fold, all the data from a single subject is removed from the training data and used to test the model. Furthermore, instead of using all the available data in the fold, we randomly selected only 10\% and 20\% of the data for training, respectively. We then train a classifier only on real data and a classifier trained on a combination of real data and synthetic data generated from the limited training data. 
In both the 10\% training data and 20\% training data scenarios, we find that combining the real and synthetic training datasets to increase the amount of data available during training significantly improved the classifier's performance $(P < 0.001)$ as seen in Figure \ref{fig:small_dataset_loso}. We observed a 27\% increase in the f1 score using motions generated by the Attention WGAN-GP model and a 23\% increase using motions generated by the LSTM in the 10\% training data scenario. In the 20\% training data scenario, we observed a lower increase of 9\% in both cases using motions generated with the Attention WGAN-GP model and by the LSTM. Thus, introducing synthetic motions into the training data in scenarios where the classifier has little real data to learn from can significantly improve performance. 

However, we did not observe any significant changes in the classification performance when we used all the available training data and doubled them with synthetic data. This is shown in Table \ref{Tab:ActionC} which demonstrates the results from both a LOSO evaluation protocol as well as Stratified K-Fold. In Stratified K-Fold ($K=10$), all the data in each fold is equally distributed across classes. Each fold has data for all subjects in this scenario, with an equal number of gait cycles for each subject affect pair. 

Nevertheless, the Attention WGAN-GP model outperformed the LSTM model in all scenarios in action classification tasks when it came to training models only on synthetic data with a very significant margin $(P < 0.001)$. We believe that introducing the generative control factor into the Attention WGAN-GP model is why the data generated by our model performs on a similar level to real data and outperforms a traditional LSTM model.

\begin{table}[!h]
\centering
\caption{Action Classification F1 scores (full dataset) }\label{tab:exp1-action-f1}
\setlength\tabcolsep{3pt}
\renewcommand{\arraystretch}{1.5}
\begin{tabular}{cccccc}
\hline
\multirow{2}{*}{Model} & \multicolumn{1}{c}{\textbf{Knock}} & \multicolumn{1}{c}{\textbf{Lift}} & \multicolumn{1}{c}{\textbf{Throw}} & \multicolumn{1}{c}{\textbf{Walk}} \\ \cline{2-6} 
 & F1  & F1 & F1 & F1 \\ \hline
\textbf{LOSO Classification} &  &  &  &    \\ \hline
Real & $92 \pm 7$ & $83 \pm 14$ & $81 \pm 11$ & $99 \pm 3$ \\
Real + Attention WGAN-GP & $92 \pm 8$ & $85 \pm 9$ & $79 \pm 15$ & $99 \pm 3$ \\
Real + LSTM & $91 \pm 8$ & $85 \pm 10$ & $81 \pm 9$ & $100 \pm 1$ \\
Attention WGAN-GP & $90 \pm 8$ & $81 \pm 11$ & $77 \pm 12$ & $99 \pm 3$ \\
LSTM & $81 \pm 11$ & $73 \pm 8$ & $50 \pm 15$ & $ 97 \pm 5$ \\ \hline
\textbf{Stratified K-Fold Classification} &  &  &  &  &  \\ \hline
Real & $97 \pm 1$ & $95 \pm 2$ & $94 \pm 2$ & $100 \pm 0$ \\
Real + Attention WGAN-GP & $97 \pm 1$ & $95 \pm 1$ & $94 \pm 2$ & $100 \pm 0$ \\
Real + LSTM & $97 \pm 1$ & $95 \pm 1$ & $93 \pm 2$ & $100 \pm 0$ \\
Attention WGAN-GP & $97 \pm 1$ & $94 \pm 2$ & $91 \pm 3$ & $100 \pm 0$ \\
LSTM & $93 \pm 2$ & $82 \pm 2$ & $68 \pm 3$ & $99 \pm 1$ \\ \hline
\end{tabular}
\label{Tab:ActionC}
\end{table}

\section{Discussion and Conclusions}
We have presented a novel autoregressive probabilistic and adversarial deep learning model based on end-to-end-training capable of both short- and long-term motion prediction through the use of attention. We have shown that this model outperforms existing recurrent LSTM commonly used models in the task of human motion synthesis, especially in generating plausible motions over long time horizons. Furthermore, we demonstrated the improved controllability over the motion synthesis of our model by creating synthetic motion datasets on which we train CNN classification models. Finally, we exploit the synthetic motion generation as a data augmentation step to improve the classification performance of deep learning models trained on real motion-captured data in scenarios where training data is lacking. 

This model can be also extended to incorporate a multi-modal control input. Although we use the control input reasonably simply for conditioning the motion class, we adopted an architecture design to allow incorporating more complex signals, such as sound and music. For example, this would allow generating complex motions, such as dancing driven by musical context. Nevertheless, to accomplish this would require expanding the architecture with additional encoding modules such that the model can incorporate information from the different modalities. Furthermore, it would require to adopt the attention layer accordingly. 

Currently, the evaluation of state-of-the-art human motion synthesis models is often based on subjective human observers reports, since it targets the domain of human computer interaction \cite{henter_moglow_2020}. Therefore, it is unclear how successful these methods are in data augmentation scenarios. In particular, it is not clear if they would overfit to particular subject characteristics. 
We ventured into this investigation, hoping that through the creation of synthetic data, we would potentially be able to improve on the inter-subject generalisability of classification of human motion. This is a particularly challenging machine learning problem since there is large variability between subjects, such as different skeleton sizes, differing positions of joints, and differences in styles of motions. 
In action classification tasks motions such as lifting, knocking, throwing and walking are performed relatively in a similar way across subjects. Therefore, we expected that the synthesis of data would enhance classification performance. Although, we observed this boost when we used a small subset of the available datasets for training, the performance does not improve significantly when using all the available training data. Further work with data from simulation environments could shed light on how data augmentation processes can become more systematic and efficient.     \\

\bibliographystyle{abbrv}
\bibliography{bibliography}


\end{document}